\documentclass[a4paper,fleqn]{cas-sc}

\usepackage[authoryear,longnamesfirst]{natbib}
\usepackage{amsthm}
\usepackage{algorithm}
\usepackage{algorithmic}
\usepackage[switch]{lineno}
\usepackage{xcolor} 
\usepackage{amsmath,amssymb,amsfonts}
\usepackage{graphicx}
\usepackage{tabularx}
\usepackage{booktabs}
\usepackage{multirow}
\usepackage{rotating}
\usepackage{float} 
\usepackage{comment}
\usepackage{subcaption}
\usepackage{nicematrix}
\newcommand{\abb}{\texttt{UniFault}\xspace}

\begin{document}
\shortauthors{E. Eldele et~al.}
\shorttitle{UniFault Foundation Model}

\title [mode = title]{UniFault: A Fault Diagnosis Foundation Model from Bearing Data}

\author[1,2]{Emadeldeen Eldele}[orcid=0000-0002-9282-0991]
\ead{emad0002@ntu.edu.sg}

\author[4]{Mohamed Ragab}
\ead{mohamedr002@e.ntu.edu.sg}

\author[2]{Xu Qing}
\ead{xu_qing@i2r.a-star.edu.sg}

\author[2]{Edward}
\ead{edward.mononym@proton.me}

\author[2,3]{Zhenghua Chen}
\ead{chen0832@e.ntu.edu.sg}

\author[2]{Min Wu}
\cormark[1]
\ead{wumin@i2r.a-star.edu.sg}

\author[2,3,5]{Xiaoli Li}
\ead{xlli@i2r.a-star.edu.sg}

\author[6]{Jay Lee}
\ead{leejay@umd.edu}

\affiliation[1]{organization={Department of Computer Science, Khalifa University},
                city={Abu Dhabi},
                postcode={127788}, 
                country={UAE}}
                
\affiliation[2]{organization={Institute for Infocomm Research, A*STAR},
                city={Singapore},
                postcode={138632}, 
                country={Singapore}}

\affiliation[3]{organization={Centre for Frontier AI Research, A*STAR},
                city={Singapore},
                postcode={138632}, 
                country={Singapore}}
                
\affiliation[4]{organization={Propulsion and Space Research Center, Technology Innovation Institute},
                city={Abu Dhabi},
                postcode={9639}, 
                country={UAE}}

\affiliation[5]{organization={College of Computing and Data Science, Nanyang Technological University},
                city={Singapore},
                postcode={639798}, 
                country={Singapore}}

\affiliation[6]{organization={Center for Industrial Artificial Intelligence, Department of Mechanical Engineering, A. James Clark School of Engineering, University of Maryland},
                city={Maryland},
                postcode={20742}, 
                country={United States of America}}                

\cortext[cor1]{Corresponding author}

\begin{abstract} 
Machine fault diagnosis (FD) is a critical task for predictive maintenance, enabling early fault detection and preventing unexpected failures.
Despite its importance, existing FD models are operation-specific with limited generalization across diverse datasets.
Foundation models (FM) have demonstrated remarkable potential in both visual and language domains, achieving impressive generalization capabilities even with minimal data through few-shot or zero-shot learning.
However, translating these advances to FD presents unique hurdles. Unlike the large-scale, cohesive datasets available for images and text, FD datasets are typically smaller and more heterogeneous, with significant variations in sampling frequencies and the number of channels across different systems and applications. This heterogeneity complicates the design of a universal architecture capable of effectively processing such diverse data while maintaining robust feature extraction and learning capabilities.
In this paper, we introduce \abb, a foundation model for fault diagnosis that systematically addresses these issues. Specifically, the model incorporates a comprehensive data harmonization pipeline featuring two key innovations. First, a unification scheme transforms multivariate inputs into standardized univariate sequences. Second, a novel cross-domain temporal fusion strategy mitigates distribution shifts and enriches sample diversity and count, improving the model generalization across varying conditions. \abb is pretrained on over 6.9 million samples spanning diverse FD datasets, enabling superior few-shot performance. Extensive experiments on real-world FD datasets demonstrate that \abb achieves state-of-the-art performance, setting a new benchmark for fault diagnosis models and paving the way for more scalable and robust predictive maintenance solutions.
\end{abstract}

\begin{keywords}
Fault Diagnosis \sep Foundation Model \sep Time Series \sep Few-shot Learning \sep Contrastive Learning \sep Transformer
\end{keywords}

\maketitle

\begin{figure*}
    \centering
    \includegraphics[width=0.9\linewidth]{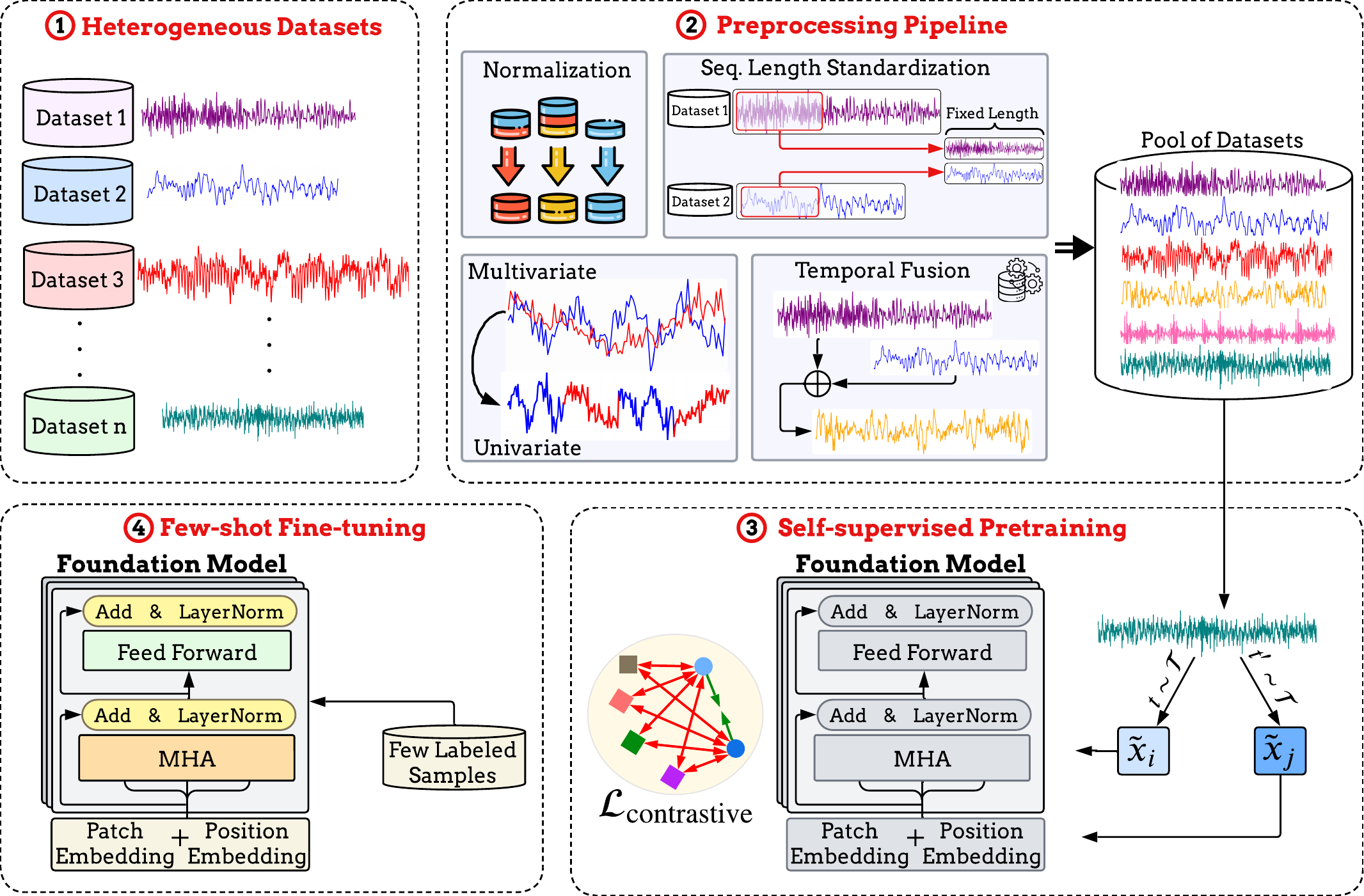}
    \caption{The overall design of \abb. (1) We collect datasets from multiple heterogeneous sources with different sequence lengths, sampling rates, and channel counts. (2) Our preprocessing pipeline includes data normalization, sequence length standardization, unifying the channel dimension, and generating new samples via cross-domain temporal fusion. (3) We perform contrastive-based self-supervised pretraining to our Transformer-based backbone. (4) The pretrained model can be fine-tuned with few-shot samples.}
    \label{fig:fugue_overall}
\end{figure*}

\section{Introduction}
Machine Fault Diagnosis (FD) plays a crucial role in predictive maintenance by ensuring the reliability and efficiency of industrial systems \cite{pr11020369}. As industries increasingly adopt automation and cost-effective operations, the demand for scalable and robust FD solutions has grown \cite{FD2015survey}. Recent advances in deep learning have revolutionized FD by enabling the automated extraction of complex patterns from sensor data, thereby detecting subtle fault signatures that often elude traditional statistical and rule-based methods \cite{DL_FD_survey_2019,s23115299,8651897}.
For instance, \cite{SINITSIN2022109454} combined convolutional neural networks (CNN) and multi-layer perceptrons (MLP), while \cite{XU2023110609, LI2025112025} used convolutional graph neural networks to process sensory bearing data and rotating machines.

Despite these achievements, significant challenges remain. Many deep learning models are highly operation-specific, struggling to generalize across diverse datasets. For instance, slight variations in sensor calibration or changes in operating conditions can lead to considerable performance degradation \cite{TL_FD_survey_2023,guo2024domain,DA_FD_survey_2021,DA_FD_survey_2024}. Furthermore, these methods typically depend on large annotated datasets—a critical limitation in real-world FD where faults are rare and manual annotation is both time-consuming and costly \cite{fink2020potential,rombach2023fault}. Such challenges underscore the need for models that can generalize effectively even with limited labeled data.

In response to these challenges, foundation models (FMs) have emerged as a transformative technology in computer vision and natural language processing. By pretraining on large-scale, heterogeneous datasets, FMs learn powerful and flexible representations that transfer effectively to downstream tasks—even when labeled data is scarce \cite{schneider2024foundation,yuan2023power}. This remarkable generalization capability makes them promising candidates for addressing the data scarcity and domain variability issues in FD.

Nonetheless, applying FMs to FD is not straightforward. Two major barriers must be overcome: (1) data scale—FD datasets are typically small and fragmented, lacking the volume required for conventional FM training \cite{su2024machine}; and (2) data heterogeneity—variations in sensor configurations, data structures, sampling rates, and other system-specific factors pose additional challenges \cite{s20247065,HUANG202154}. 
Another critical obstacle is the presence of \textbf{distribution shifts} across datasets. Vibration signals collected from different machines, sensors, or operating conditions often follow distinct statistical distributions, even when they represent the same fault type. For example, the spectral profile of an outer-race defect may vary significantly between datasets due to differences in sensor placement, load levels, or sampling hardware. Such distributional mismatches hinder the transferability of features learned on one dataset to another, limiting the robustness of existing FD models.
Our work aims to tackle these obstacles by exploring novel approaches for adapting FMs to the unique demands of fault diagnosis.

In this paper, we systematically address the aforementioned challenges by proposing a \textbf{Uni}fied foundation model for bearing \textbf{Fault} diagnosis (\textbf{\abb}), which introduces three key contributions. First, to overcome the scarcity of large annotated datasets, we have constructed a large-scale, diverse FD database comprising over \textit{6.9 million} samples collected from heterogeneous sources. \abb leverages this extensive dataset for pretraining, allowing the model to learn generalized representations across varied operating conditions. Second, to tackle the issue of data heterogeneity, we develop a comprehensive data harmonization pipeline. This pipeline features a channel unification scheme that converts diverse multivariate sensor inputs into univariate sequences while retaining local inter‐channel relationships. Moreover, a cross-dataset temporal fusion strategy is integrated to mitigate distribution shifts and enrich sample diversity, thereby enhancing both robustness and generalization.

Unlike existing FD models that are typically narrow in scope or require manual adaptation across datasets, \abb addresses the core challenges of data scarcity, heterogeneity, and the absence of a general-purpose architecture in FD, laying the groundwork for a scalable, robust, and universal solution. 

We further validate \abb through extensive fine-tuning experiments, demonstrating its remarkable ability to achieve high performance with limited labeled data—even with as few as 100 samples. This strong few-shot learning capability positions \abb as an effective foundation model for real-world FD applications, particularly in scenarios where labeled data is limited.

In summary, the key contributions of this work are as follows:

\begin{itemize}
    \item We introduce \abb, a general-purpose foundation model for fault diagnosis pretrained on over 6.9 million samples, significantly surpassing the scale of any prior FD models, to enable generalization across datasets, domains, and machine types.
    \item We present a systematic preprocessing pipeline that standardizes heterogeneous datasets via a normalization scheme and enhances robustness with a cross-domain temporal fusion strategy.
    \item We conduct extensive fine-tuning experiments on real-world FD datasets, demonstrating that \abb exhibits remarkable few-shot learning performance and benefits significantly from our preprocessing pipeline.
\end{itemize}

The remainder of this paper is organized as follows: Section \ref{sec:related} provides an overview of related work in fault diagnosis and foundation models. Section \ref{sec:methods} details the data preprocessing pipeline, our model architecture, and the self-supervised training strategy. Section \ref{sec:exp_setup} presents the details of the datasets, the experimental setup, and baselines. Section \ref{sec:res} shows the evaluation results and some key experiments. Finally, Section \ref{sec:conc} concludes the paper.

\section{Related Works}
\label{sec:related}
\subsection{Deep Learning for Fault Diagnosis}

Deep learning has significantly advanced fault diagnosis (FD) by enabling automated feature extraction and capturing complex temporal patterns. Convolutional Neural Networks (CNNs) have been widely used to extract discriminative features from sensor data \cite{CNN_FD_review1,CNN_FD1,CNN_FD2}, while Long Short-Term Memory networks (LSTMs) effectively model sequential dependencies \cite{LSTM_FD,LSTM_FD2}. More recent approaches have adapted Transformer architectures \cite{transformer_FD1,transformer_FD2,transformer_FD3} and Temporal Convolutional Networks (TCNs) \cite{tcn} to further improve pattern recognition and robustness under varying conditions.

However, many of these methods are tailored to specific domains or tasks, often assuming that training and testing data come from the same distribution. In practice, variations in sensor configurations, sampling rates, and operating conditions limit their generalizability. Moreover, the reliance on large, labeled datasets—which are frequently unavailable in industrial settings—further impedes their scalability and practical deployment.

\subsection{Foundation Models for Time Series}
Recently, large-scale foundation models such as GPT-4.5 \cite{achiam2023gpt}, Gemini \cite{team2024gemini}, and DeepSeek \cite{liu2024deepseek} have transformed domains like natural language processing and computer vision through advancements in self-supervised learning and zero-shot generalization capabilities. Similar methodologies have begun to be adapted for time series analysis \cite{liang2024foundation}, with frameworks such as Time-LLM \cite{jin2023time} and UniTime \cite{liu2024unitime} tackling forecasting tasks through prompt-based strategies and domain-specific adaptations. Additionally, specialized transformer-based models like {MOMENT} \cite{goswami2024moment} leverage diverse public time series datasets to deliver versatile performance across analytical tasks, while {GPT4TS} \cite{onefitsall} extends pretrained GPT-2 models to time series domains by fine-tuning only task-specific linear layers. Convolutional approaches, exemplified by {TSLANet} \cite{tslanet}, incorporate adaptive spectral and interactive convolutional blocks, enhancing representation learning specifically for time series.

Nevertheless, these models generally assume access to extensive labeled data and often do not adequately address the inherent heterogeneity and diverse analytical requirements encountered in real-world time series datasets, particularly in fault diagnosis applications.

\subsection{Foundation Models for Fault Diagnosis}
Very recently, a few studies have begun applying foundation model concepts directly to fault diagnosis. For example, one study on electrical motor fault diagnosis employs self-supervised learning to build a robust backbone that demonstrates promising performance across different machines and operating conditions \cite{anbalagan2023foundational}. Similarly, another work in bearing fault diagnosis introduces a cloud-edge-end semi-supervised framework that, through tailored data augmentation and contrastive learning strategies, achieves high accuracy using only a small fraction of labeled data \cite{lai2024bearingfm}.

Despite these encouraging results, both studies are constrained by their reliance on relatively small, single-source datasets for pretraining and tend to overlook the challenges posed by heterogeneous sensor configurations and distribution shifts in real-world industrial environments. In contrast, our proposed Fault Diagnosis Foundation Model (\abb) addresses these limitations by leveraging a massive, heterogeneous FD dataset comprising over 6.9 million samples for pretraining. Moreover, \abb employs a unified preprocessing pipeline, including a novel cross-dataset temporal fusion strategy that generates intermediate-domain samples to expose the model to smoother distribution transitions and mitigate distribution shifts. This paradigm is akin to the domain adaptation techniques \cite{ganin2016domain,tzeng2017adversarial}), which aim to align a labeled source with an unlabeled target. Detailed discussions of our methodology and contributions are provided in Section \ref{sec:methods}.

\section{Methods} 
\label{sec:methods}
\subsection{Problem Formulation}
Let $\mathcal{X} = \{X_i\}_{i=1}^N$ denote a collection of heterogeneous fault diagnosis datasets, where each $X_i \in \mathbb{R}^{C_i \times L_i}$ represents a multivariate time series sequence. Here, $C_i$ denotes the number of channels (e.g., different sensors), and $L_i$ is the sequence length. Due to variations in sensor configurations, sampling rates, and operational conditions, both $C_i$ and $L_i$ differ across datasets, leading to heterogeneous data structures and domain shifts.

The objective is to develop a foundation model $f_\theta$, parameterized by $\theta$, that: (i) Processes heterogeneous inputs by unifying $X_i$ into a standardized representation space; (ii) Extracts robust, domain-invariant features $\mathbf{z}_i = f_\theta(X_i)$ that effectively capture the underlying patterns in diverse datasets; and (iii) Adapts to new tasks with minimal labeled data via few-shot fine-tuning. 

The overall learning process involves two stages:
\begin{enumerate}
    \item \textbf{Pretraining}: Given unlabeled or sparsely labeled datasets $\mathcal{X}$, optimize $\theta$ to minimize:
   \[
   \theta^* = \arg \min_{\theta} \mathbb{E}_{X_i \sim \mathcal{X}} [\mathcal{L}_{\text{pretrain}}(f_\theta(X_i))],
   \]
   where $\mathcal{L}_{\text{pretrain}}$ is a pretraining loss designed to learn generalized representations from diverse, potentially unlabeled datasets.

   \item \textbf{Few-Shot Fine-Tuning}: For a target task with $m \ll N$ labeled samples $\{(X_j, y_j)\}_{j=1}^m$, adapt $f_{\theta^*}$ by a minimal fine-tuning of $\theta^*$, with training a lightweight adapter $g_\phi$ to predict labels:  
\[
\phi^* = \arg \min_{\phi} \mathbb{E}_{(X_j, y_j)} \left[ \mathcal{L}_{\text{CE}}\left(y_j, \, g_\phi\left(f_{\theta^*}(X_j)\right)\right) \right],
\]  
such that $\mathcal{L}_{\text{CE}}$ is the cross-entropy loss.  
\end{enumerate}

\subsection{Overview}
The proposed \abb framework addresses the heterogeneity of fault diagnosis data through: (1) a universal data preprocessing pipeline to unify diverse FD datasets into a standardized format; and (2) the Transformer model, which processes harmonized data with a temporal self-attention mechanism optimized for machinery signals. The overall framework is illustrated in Fig.~\ref{fig:fugue_overall}.

\subsection{Data Preprocessing Pipeline}
For effective pretraining, the preprocessing pipeline addresses five major inconsistencies in fault diagnosis (FD) data: variable sampling rates, unequal sequence lengths, differing numbers of channels, mismatched feature scales, and distribution mismatches across datasets. The first four are handled through dataset splitting, resampling, length standardization, channel unification, and normalization. The final issue is addressed through a cross-domain temporal fusion strategy, which creates intermediate representation spaces among heterogeneous datasets.

\subsubsection{Train–Validation–Test Splitting}
Each dataset is first partitioned into training, validation, and test splits. This ensures that preprocessing transformations such as resampling, downsampling, and normalization are performed in a manner consistent with the training distribution, preventing information leakage into the validation and test sets.

\subsubsection{Resampling}
Raw FD datasets are collected under different acquisition rates, leading to inconsistent temporal resolutions. To resolve this, we resample all signals to a fixed temporal resolution of 0.1 seconds. This value is chosen empirically based on dataset statistics: most datasets yield sequences exceeding 1024 timesteps under this setting, ensuring sufficient temporal granularity while maintaining computational efficiency.

\subsubsection{Sequence Length Standardization}
Even after resampling, datasets produce windows of varying lengths. To unify input dimensions across datasets, we downsample each sequence to a fixed length of 1024 timesteps. This guarantees consistency for model training, while still preserving the temporal dynamics necessary for detecting faults.

\subsubsection{Channel-Invariant Unification}
Datasets often contain different numbers of sensor channels due to variations in hardware and setup. To avoid dataset-specific bias, we convert multivariate sequences into a univariate form by treating each channel independently \cite{patchtst}. This channel-invariant representation allows the model to learn generic temporal dynamics without relying on a fixed channel configuration. During fine-tuning, the original channel structure of the target dataset is preserved.

\subsubsection{Data Normalization} 
Finally, signals are normalized using statistics computed only from the training split. Each channel is scaled via min–max normalization into a fixed numerical range, ensuring stability during optimization and promoting comparability across datasets. By restricting normalization parameters to the training set, we avoid distributional bias in the validation and test sets.

\subsubsection{Cross-Domain Temporal Fusion}
To mitigate distribution shifts and enhance sample diversity across heterogeneous fault diagnosis datasets, we propose a Cross-Domain Temporal Fusion strategy inspired by \cite{cotmix} but adapted for foundation model pretraining. While the original method focused on pairwise domain adaptation, our approach generalizes to arbitrary cross-dataset interactions, enabling synthetic sample generation from any pair of pretraining datasets while learning their temporal relationships. This fosters robustness to unseen operational conditions and sensor configurations.

Given two univariate time series samples \(X_a \in \mathbb{R}^{L}\) (from dataset \(a\)) and \(X_b \in \mathbb{R}^{L}\) (from dataset \(b\)), we generate fused samples \(X_{\text{fused}}\) by choosing a dominant dataset (e.g., \(a\)). Then, for each timestep \(i\) in the fused sample, we combine \(X_a^i\) with a temporal neighborhood of \(X_b\) as follows:  
   \[
   X_{\text{fused}}^i = \lambda X_a^i + (1 - \lambda) \cdot \frac{1}{T} \sum_{j=i-T/2}^{i+T/2} X_b^j, \quad 0.5 < \lambda < 1, \tag{1}
   \]  
   where \(T\) is the temporal window size, and \(\lambda\) is kept \(> 0.5\) to control the dominance of \(X_a\).  

To ensure balanced augmentation, this process is done in a bidirectional manner. Specifically, we generate both \(a\)-dominant and \(b\)-dominant samples:  
   \[
   \begin{aligned}
   X_{\text{fused},a}^i &= \lambda X_a^i + (1-\lambda) \cdot \text{MA}(X_b, i, T), \\
   X_{\text{fused},b}^i &= \lambda X_b^i + (1-\lambda) \cdot \text{MA}(X_a, i, T),
   \end{aligned}
   \]  
where \(\text{MA}(\cdot)\) denotes a moving average over \(T\) timesteps centered at \(i\)---a process to learn the temporal information in the less dominant domain. 

During pretraining, fused samples are treated as additional training data. By exposing the model to interpolated domains, \abb learns to disentangle fault-related patterns from domain-specific variations

\subsection{Model Architecture}
At its core, \abb builds upon the Transformer architecture \cite{vit}. 
While Transformers were originally developed for natural language processing, their self-attention mechanism is particularly well-suited for vibration and machinery signals. 
Fault signatures often manifest as periodic or cyclostationary transients that recur with variable spacing due to load fluctuations or machine variability. 
Unlike CNNs, which rely on fixed receptive fields \cite{tslanet}, or RNNs, which struggle with long memory \cite{pmlr-v119-zhao20c}, Transformers can flexibly capture such long-range temporal dependencies. 
Moreover, the ability of self-attention to assign different weights to different time positions allows the model to focus on fault-induced bursts even when they are sparse or masked by noise. 
To handle long sequences efficiently, \abb employs patch embeddings and dimensionality reduction, ensuring the architecture remains scalable without an explosion in parameters. 
Next, we briefly discuss its architectural components.

\paragraph{\textbf{Input Embedding:}}
After preprocessing, the input univariate data to the transformer ($X \in \mathbb{R}^{1 \times L}$, where $L$ is the sequence length) are projected into a \(d\)-dimensional space via a linear layer, producing token embeddings \(\mathbf{E} \in \mathbb{R}^{L \times d}\).

\paragraph{\textbf{Positional Encoding:}}
Learnable positional encodings \(\mathbf{P} \in \mathbb{R}^{L \times d}\) are added to \(\mathbf{E}\) to retain temporal order:  
   \[
   \mathbf{Z}_0 = \mathbf{E} + \mathbf{P}.
   \]

\paragraph{\textbf{Transformer Layers:}}
The model stacks \(N\) identical layers, each comprising:
\begin{itemize}
    \item \textit{Multi-Head Self-Attention}: Captures global temporal dependencies.  
     \[
     \mathbf{Q}, \mathbf{K}, \mathbf{V} = \mathbf{Z}_{l-1}\mathbf{W}_Q, \mathbf{Z}_{l-1}\mathbf{W}_K, \mathbf{Z}_{l-1}\mathbf{W}_V,
     \]  
     \[
     \text{Attention}(\mathbf{Q}, \mathbf{K}, \mathbf{V}) = \text{softmax}\left(\frac{\mathbf{Q}\mathbf{K}^T}{\sqrt{d}}\right)\mathbf{V}.
     \]  
    \item \textit{Feed-Forward Network}: A two-layer MLP with GELU activation.
    \item \textit{Layer Normalization}: Applied pre-attention and pre-MLP.
\end{itemize}

\subsection{Self-Supervised Learning}
Labeled fault diagnosis datasets are typically scarce, as annotating real machinery faults is expensive and sometimes impossible. In contrast, large volumes of unlabeled vibration signals are readily available. To exploit this data, \abb is pretrained with a contrastive self-supervised learning (SSL) strategy.

With contrastive SSL, we apply augmentations to signals, e.g., add noise or shift the signal, and the model is trained to recognize that the different augmented views of the same signal should have similar representations, while signals from different machines or conditions should remain distinguishable. This way, the model learns to capture intrinsic fault-related patterns before any supervised fine-tuning.

\subsubsection{Augmentation Strategies}
We design augmentations to mimic variations commonly encountered in industrial settings:  

\textbf{Temporal Shifting}: Fault signatures (e.g., repetitive impacts from a crack) may appear at different phases depending on load or speed. By cyclically shifting the sequence, we enforce invariance to such phase differences:  
\[
X'_{\text{shift}}(t) = X\!\big((t - sL) \bmod L\big),
\]  
where $s$ is a random shift ratio and $L$ is the sequence length.  

\textbf{Scaling with Sensor Jitter}: Sensor gain, calibration, and background noise vary across machines. We simulate this with multiplicative scaling and additive noise:  
\[
X'_{\text{scale}} = X \odot \mathbf{F} + \mathbf{J}, \quad \mathbf{F}\!\sim\!\mathcal{D}_F,\; \mathbf{J}\!\sim\!\mathcal{D}_J,
\]  
ensuring robustness to amplitude changes and high-frequency disturbances.  

These augmentations directly address FD challenges by promoting invariance to operating condition shifts and sensor variability.  

\subsubsection{Contrastive Loss}
For each sample $X_i$, two augmented views $(X'_i, X''_i)$ are encoded into embeddings $(\mathbf{z}'_i, \mathbf{z}''_i)$. We maximize their similarity while contrasting them against other signals in the batch:  
\[
\mathcal{L}_{\mathrm{cont}} = -\tfrac{1}{N}\sum_{i=1}^{N}\log\frac{A_{ii}}{\sum_{k=1}^{N}(A_{ik}+B_{ik})},
\]  
with  
\[
A_{ik} = \exp\!\left(\tfrac{\mathrm{sim}(\mathbf{z}'_i,\mathbf{z}''_k)}{\tau}\right), \quad 
B_{ik} = \exp\!\left(\tfrac{\mathrm{sim}(\mathbf{z}'_i,\mathbf{z}'_k)}{\tau}\right).
\]  

Here, $\mathrm{sim}(\cdot)$ denotes cosine similarity and $\tau$ is the \textbf{temperature} hyperparameter. In the FD context, $\tau$ controls how tightly the model clusters fault signatures, such that a smaller $\tau$ emphasizes fine-grained distinctions (useful for early fault detection), while a larger $\tau$ encourages broader invariance across operating conditions. By balancing these effects, \abb learns representations that are both sensitive to subtle fault patterns and robust across machines and environments.

\section{Experimental Settings}
\label{sec:exp_setup}
This section describes the datasets, model variants, our experimental setup, and the baselines. These details are essential for replicating the experiments and validating the generalizability of the proposed model.

\subsection{Datasets} 

\begin{table*}[h]
    \centering
    \caption{Summary of bearing fault diagnosis datasets used in this study.}
    \resizebox{\textwidth}{!}{
    \begin{NiceTabular}{l|p{4cm}p{7.5cm}cp{2cm}}
        \toprule
        \textbf{Dataset} & \textbf{Fault Generation} & \textbf{Operating Conditions} & \textbf{NOConditions} & \textbf{SRate (Hz)} \\
        \midrule
        \textbf{IMS} & Naturally degrading over time & Constant speed of 2000 RPM with a 6000 lbs radial load & 1 & 20,000 \\ \midrule
        \textbf{UO} & Artificial faults induced by EDM & Various operating conditions, including different load levels and rotational speeds & Multiple & 42,000 \\ \midrule
        \textbf{CWRU} & Artificially induced & Motor loads (0–3 HP, 1797–1720 RPM) with faults in inner raceway, rolling element, and outer raceway at 6, 3, and 12 o’clock & Multiple & 12,000 or 48,000 \\ \midrule
        \textbf{PU} & Artificial and real damages & Four different operating conditions: varying rotational speed, load torque, and radial force & 4 & 64,000 \\ \midrule
        \textbf{Torino} & Artificially induced (0–450 µm) & Speeds (0–500 Hz), loads (0–1800 N) & Multiple & 25,600 \\ \midrule
        \textbf{XJTU-SY} & Natural degradation via accelerated life testing & Three conditions: 2100 rpm (12 kN), 2250 rpm (11 kN), 2400 rpm (10 kN) & 3 & 25,600 \\ \midrule
        \textbf{MFPT} & Artificially induced faults & Various operating conditions, including different load levels and rotational speeds & Multiple & 97,656 \\ \midrule
        \textbf{FEMTO} & Natural degradation via accelerated life testing & Variable speeds and loads & 3 & 25,600 \\ \midrule
        \textbf{KAIST} & Natural degradation via accelerated life testing & Constant 1770–1780 RPM with axial load (2.94 kN) and vertical load (5.88 kN) & 1 & 25,600 \\ \midrule
        \textbf{HIT-SM} & Artificially induced faults using EDM & Three speeds (600, 900, and 1200 RPM) with different radial loads & 3 & 51,200 \\
        \midrule
        \textbf{CNC} & Naturally occurring and operational faults & Three brownfield milling machines under varying tool operations and production loads & Multiple & 2,000 \\
        \bottomrule
    \end{NiceTabular}
    }
    \label{tab:dataset_summary}
\end{table*}

To develop a robust foundation model for fault diagnosis, we pretrain \abb on a large and diverse collection of ten bearing datasets, ensuring coverage of a wide range of fault types, machine configurations, and operating conditions. This extensive dataset collection enables the model to learn generalizable representations across different fault domains. For fine-tuning and evaluation, we select a subset of three representative datasets—IMS, UO, and PU—that exhibit distinct characteristics in terms of data distribution, sensor setup, and operational environments. The fine-tuning samples are excluded from the pretraining data. A summary of each dataset is described next.

The \textbf{Center for Intelligent Maintenance Systems (IMS)} dataset \cite{IMSdataset} is widely used for benchmarking fault diagnosis models. This dataset comprises run-to-failure experiments conducted under consistent operating conditions, capturing the natural progression of bearing defects over time. It contains three types of faults developed naturally over extended operational periods. Since IMS is normally designed for remaining useful life prediction, we utilized it for fault diagnosis by segmenting the vibration data into predefined health states based on timestamps\footnote{Check \url{https://github.com/Miltos-90/Failure_Classification_of_Bearings}.}.

The \textbf{Case Western Reserve University (CWRU)} dataset \cite{CWRUdataset} is another widely used dataset that features artificially induced bearing faults. It provides precise control over fault locations and sizes (inner race, outer race, and rolling element defects), allowing for a structured evaluation of model performance across known fault types.

The \textbf{Paderborn University (PU)} dataset \cite{PUdataset} includes both artificially damaged and naturally degraded bearings operating under different load conditions, making it particularly useful for assessing the robustness against domain shifts. PU provides a more diverse representation of fault progression, enhancing the adaptability of the foundation model to real-world degradation patterns.

The \textbf{IEEE PHM 2012 (FEMTO)} dataset \cite{FEMTOdataset}, developed for a prognostics competition, includes highly detailed vibration data from progressively degraded bearings. This dataset is instrumental in training \abb to recognize subtle fault progression and predict potential failures before they become critical. 

The \textbf{Xi’an Jiaotong University, Changxing Sumyoung Technology (XJTU-SY)} dataset \cite{XJTU-SYdataset} similarly captures fault progression under diverse load conditions, offering additional variability in machine operating states. The dataset contains long-term recordings of bearing degradation, which enables the study of remaining useful life estimation and the transition from healthy to faulty states.

The \textbf{MFPT bearing} dataset \cite{mfpt_dataset}, provided by the Machinery Failure Prevention Technology Society, includes vibration data from multiple fault modes recorded with high-resolution accelerometers. This dataset is valuable for studying different fault types under controlled conditions, making it an important component in the pretraining phase. 

The \textbf{KAIST ball bearing} vibration dataset \cite{kaist_dataset} was collected at the Korea Advanced Institute of Science and Technology (KAIST). This dataset focuses on high-speed bearings operating under different load and lubrication conditions, providing an additional challenge for the model to generalize across different machine settings. 

The \textbf{HIT-SM bearing} dataset \cite{hitsm_dataset}, provided by the Sensing and Measurement Laboratory at Harbin Institute of Technology, includes both healthy and faulty bearings operating at variable speeds. The dataset’s diversity in speed and load conditions further enhances the ability of our model to generalize across a wide range of machinery setups.

The \textbf{Bosch CNC Machining} dataset \cite{CNCdataset} provides vibration data collected from three brownfield milling machines operating in an actual industrial environment over a two-year period. Those are named as M01, M02, and M03. The dataset captures real-world machining variability across different machines, time intervals, and tool operations, reflecting realistic challenges such as tool wear, misalignment, and chip clogging. Data were gathered using low-cost triaxial accelerometers mounted on the spindle housing, offering non-intrusive monitoring of 15 distinct tool operations.

Lastly, the Politecnico di Torino (\textbf{Torino}) dataset \cite{DAGA2019252} consists of vibration signals collected from a high-speed spindle test rig equipped with roller bearings under controlled operating conditions. It includes both stationary measurements across varying speeds (0–500 Hz) and loads (0–1800 N), as well as endurance tests capturing the progression of faults over time. The dataset features artificially induced faults (indentations on rollers and inner rings) with different severities (0–450 µm), making it valuable for fault detection, classification, and progression analysis.

\subsection{Model Variants}
To accommodate different resource constraints and application needs, we train two variants of our \abb model, i.e., Lite and Base. These variants have different hidden dimensions, different numbers of Transformer layers, and different numbers of attention heads, as described in Table~\ref{tbl:variants_descr}.

\begin{table}[h]
    \centering
    \caption{Description of the three variants of our model.}
    \label{tbl:variants_descr}
    \begin{NiceTabular}{l|cccc}
        \toprule
        Model & Hidden Dim. & Layers & Heads & Parameters \\
        \midrule
        \abb-Lite & 128 & 4 & 4 & 823K \\
        \abb-Base  & 256 & 8 & 8 & 6.4M \\
        \bottomrule
    \end{NiceTabular}
    
\end{table}

\subsection{Training Protocol}
Both training and pretraining were performed with AdamW optimizer (\(\beta_1=0.9, \beta_2=0.95\)), learning rate \(1 \times 10^{-3}\), weight decay of \(1 \times 10^{-5}\). A cosine learning rate scheduler with warm restarts is employed to control the learning rate during training.
For pretraining, we used a batch size of 512 and trained the model for 5 epochs. For fine-tuning, we used a smaller batch size of 64 and trained the model for 200 epochs.
The experiments were conducted on NVIDIA L40 GPU using mixed-precision training. All experiments were repeated 3 times, where we report the mean ± std.

For the \textit{Few-Shot protocol}, we fine-tuned with \textit{randomly selected} 100 samples in the IMS dataset, and 1\% of data in PU and CNC datasets. Notably, we did not include any samples from these three datasets in the pretraining.

\subsection{Baselines}
\label{sec:baselines}
We compare \abb against the following three categories of state-of-the-art baselines in our experiments.

\textbf{Fault Diagnosis Models.} This category includes \textbf{C-Trans} \cite{c-trans}, \textbf{WDCNN} \cite{wdcnn}, \textbf{QCNN} \cite{qcnn} and \textbf{EverAdapt} \cite{EDWARD2025112317}. C-Trans \cite{c-trans} integrates CNNs with transformers to improve fault diagnosis in rotating machinery across various operating conditions. WDCNN \cite{wdcnn} is a deep CNN utilizing wide first-layer kernels to process raw vibration data for fault diagnosis. QCNN \cite{qcnn} is a CNN utilizing quadratic neurons to enhance feature representation and interpretability in bearing fault diagnosis. EverAdapt \cite{EDWARD2025112317} features continual batch normalization and class-conditional domain alignment to enable continuous model adaptation in dynamic environments. 

\textbf{Time Series Representation Learning Methods.} This category includes \textbf{TS2VEC} \cite{ts2vec}, \textbf{TS-TCC} \cite{tstcc} and \textbf{ROCKET} \cite{ROCKET}. TS2VEC \cite{ts2vec} employs hierarchical contrastive learning over augmented context views to derive robust, multi-scale representations for time series data. TS-TCC \cite{tstcc} utilizes weak and strong augmentations alongside novel temporal and contextual contrasting modules to learn robust and discriminative representations from unlabeled time-series data. ROCKET \cite{ROCKET} transforms time series data using a large number of random convolutional kernels and employs the resulting features to train a linear classifier.

\textbf{Time Series Foundation Models.}  This category includes \textbf{MOMENT} \cite{goswami2024moment}, \textbf{GPT4TS} \cite{onefitsall}  and \textbf{TSLANet} \cite{tslanet}.  MOMENT \cite{goswami2024moment} is a transformer-based model designed for versatile time series analysis tasks and is pretrained on a diverse collection of public time series data to enhance performance across various applications. We included the base variant of MOMENT since it achieved the best performance over the other variants. GPT4TS \cite{onefitsall} leverages the frozen pretrained GPT2 model for time series analysis and fine-tunes a linear layer for different tasks. We kept the default settings of using 6 GPT layers. TSLANet \cite{tslanet} is a convolutional model leveraging adaptive spectral and interactive convolution blocks to improve time series representation learning across multiple tasks.

Notably, the `Fault Diagnosis Models' were trained in a fully supervised manner on the few-shot datasets, without pretraining. The `Time Series Representation Learning Methods' were pretrained and then fine-tuned with the few-shot samples. Last, the `Time Series Foundation Models' are already pretrained, so we just fine-tune them directly with the few-shot samples. We follow the same training protocol for the three categories.

\begin{table*}[h]
    \centering
    \caption{Few-shot supervised fine-tuning results for IMS, PU, and M0x datasets. Baselines are categorized into Time Series Representation Learning methods (TS RL), Fault Diagnosis Models (FD), and Time Series Foundation Models (TS FM). The superscript \textsuperscript{T} indicates trainable parameters, while \textsuperscript{N} indicates non-trainable parameters (parameters are trainable by default).}
    \label{tbl:sota_comparison}
    \resizebox{\linewidth}{!}{
    \begin{NiceTabular}{l l|c|cc|cc|cc|cc|cc}
        \toprule
        \multirow{2}{*}{} & \multirow{2}{*}{\textbf{Method}} & \multirow{2}{*}{\textbf{\#Parameters}} 
        & \multicolumn{2}{c}{\textbf{IMS}} 
        & \multicolumn{2}{c}{\textbf{PU}} 
        & \multicolumn{2}{c}{\textbf{M01}} 
        & \multicolumn{2}{c}{\textbf{M02}} 
        & \multicolumn{2}{c}{\textbf{M03}} \\
        \cmidrule(lr){4-5} \cmidrule(lr){6-7} \cmidrule(lr){8-9} \cmidrule(lr){10-11} \cmidrule(lr){12-13}
        & & & ACC & F1 & ACC & F1 & ACC & F1 & ACC & F1 & ACC & F1 \\
        \midrule
        \multirow{4}{*}{\begin{turn}{90}FD\end{turn}}  
        & C-Trans  & 865 K & 55.8 ± 9.1 & 44.3 ± 9.4 & 41.2 ± 4.8 & 39.2 ± 4.7 & 96.5 ± 0.2 & 74.0 ± 2.5 & 95.4 ± 2.6 & 63.5 ± 4.2 & 99.0 ± 0.1 & 49.6 ± 0.1 \\ 
        & WDCNN  & 76.8 K & 71.0 ± 4.6 & 62.0 ± 8.4 & 29.4 ± 1.2 & 28.3 ± 2.1 & 95.6 ± 0.5 & 62.4 ± 8.9 & 96.8 ± 0.3 & 54.4 ± 0.7 & 99.0 ± 0.0 & 49.7 ± 0.0 \\
        & QCNN  & 171 K & 53.8 ± 5.8 & 52.3 ± 5.6 & 35.6 ± 11.6 & 33.6 ± 13.2 & 96.5 ± 0.5 & 73.6 ± 6.1 & 97.4 ± 0.1 & 70.6 ± 2.0 & 98.9 ± 0.2 & 57.9 ± 3.7 \\
        & EverAdapt & 200.1K & 81.2 ± 1.4 & 78.5 ± 2.9 & 72.0 ± 0.7 & 71.0 ± 1.0 & 96.8 ± 0.4 & 76.5 ± 2.8 & 97.2 ± 0.2 & 68.5 ± 1.5 & 99.1 ± 0.1 & 58.5 ± 5.2 \\
        \midrule
        \multirow{3}{*}{\begin{turn}{90}TS RL\end{turn}}  
        & TS2VEC  & 637.3 K & 88.9 ± 0.2 & 87.6 ± 1.4 & 68.4 ± 1.1 & 68.4 ± 0.9 & 95.9 ± 0.1 & 49.2 ± 0.0 & 96.8 ± 0.2 & 49.6 ± 0.0 & 98.8 ± 0.1 & 49.9 ± 0.0 \\
        & TS-TCC  & 256 K & 78.5 ± 1.8 & 70.2 ± 2.5 & 72.8 ± 2.1 & 72.5 ± 2.3 & 96.2 ± 0.3 & 65.8 ± 4.2 & 96.9 ± 0.2 & 62.3 ± 2.1 & 99.0 ± 0.1 & 52.4 ± 2.8 \\
        & ROCKET  & 0\textsuperscript{T}, 100.1K\textsuperscript{N} & 77.1 ± 0.9 & 66.0 ± 0.4 & \textbf{80.1 ± 0.2} & \textbf{80.3 ± 0.2} & 97.3 ± 0.1 & 81.4 ± 0.6 & \textbf{98.0 ± 0.0} & \textbf{77.7 ± 0.3}  & 99.0 ± 0.0 & 49.7 ± 0.0 \\
        \midrule
        \multirow{3}{*}{\begin{turn}{90}TS FM\end{turn}}  
        & MOMENT-Base  & 3.1K\textsuperscript{T}, 109.6M\textsuperscript{N} & 82.4 ± 1.5 & 74.6 ± 1.8 & 54.2 ± 1.2 & 53.8 ± 1.4 & 96.4 ± 0.3 & 68.2 ± 3.5 & 96.7 ± 0.3 & 60.8 ± 2.2 & 99.0 ± 0.1 & 50.5 ± 1.5 \\
        & GPT4TS & 1.3M\textsuperscript{T}, 81.1M\textsuperscript{N} & 36.2 ± 1.1 & 20.4 ± 0.9 & 38.5 ± 1.5 & 36.2 ± 1.8 & 95.2 ± 0.8 & 58.4 ± 5.6 & 95.8 ± 0.5 & 55.2 ± 3.4 & 98.9 ± 0.1 & 49.8 ± 0.5 \\
        & TSLANet  & 531 K & 91.3 ± 0.5 & 83.1 ± 0.9 & 69.8 ± 2.4 & 69.6 ± 2.8 & 54.1 ± 7.1 & 42.0 ± 7.8 & 95.4 ± 2.0 & 71.0 ± 2.9  & 99.0 ± 0.0 & 49.7 ± 0.0 \\
        \midrule
        \multirow{2}{*}{\begin{turn}{90}Ours\end{turn}}  
                & \abb-Lite  & 823 K & \underline{95.5 ± 1.6 }& \underline{96.2 ± 1.3} & 76.4 ± 1.8 & 76.1 ± 2.1 & \underline{97.5 ± 0.1} & \underline{82.9 ± 1.2} & 97.6 ± 0.0 & 70.1 ± 0.7 & \underline{99.2 ± 0.1} & \underline{63.8 ± 7.7} \\
        & \abb-Base  & 6.4 M & 
        \textbf{98.1 ± 1.0} & \textbf{98.3 ± 0.8} & \underline{79.3 ± 0.5} & \underline{79.4 ± 0.4} & \textbf{97.6 ± 0.2} & \textbf{84.0 ± 1.4} & \underline{97.7 ± 0.0} & \underline{71.2 ± 0.5} & \textbf{99.2 ± 0.1} & \textbf{64.8 ± 8.7} \\
        \bottomrule
    \end{NiceTabular}
    }
\end{table*}

\section{Results}
\label{sec:res}

Table~\ref{tbl:sota_comparison} presents the fine-tuning results on the IMS, UO, and CNC datasets, comparing our proposed \abb model variants (Lite and Base) against baseline methods from the three categories described in Section~\ref{sec:baselines}.
The results show that the proposed \abb models demonstrate strong and consistent few-shot fine-tuning performance across all datasets. On the challenging IMS dataset, \abb-Base achieves the highest accuracy and F1-score (98.1\% / 98.3\%), outperforming all Time-Series Representation Learning and Fault Diagnosis baselines. The smaller \abb-Lite variant also delivers competitive results (95.5\% / 96.2\%), confirming that efficient fine-tuning can be achieved without compromising representational robustness. These findings underscore the effectiveness of hierarchical Transformer-based pretraining in modeling complex, evolving fault dynamics.

Across the PU and CNC datasets, \abb-Base remains among the top-performing methods, matching or narrowly trailing ROCKET on specific subsets while surpassing all FD and TS FM baselines in overall stability. The \abb variants maintain high accuracy and superior F1-scores, particularly on complex component datasets (M01, M03), highlighting their adaptability to both naturally evolving and synthetic fault domains. Overall, large-scale time-series pretraining enables \abb to generalize effectively under few-shot supervision, outperforming traditional models that depend on handcrafted or shallow representations.

\subsection{Ablation Study}
\subsubsection{Effect of Cross-dataset Temporal Fusion}

\begin{figure}
    \centering
    \includegraphics[width=0.45\columnwidth]{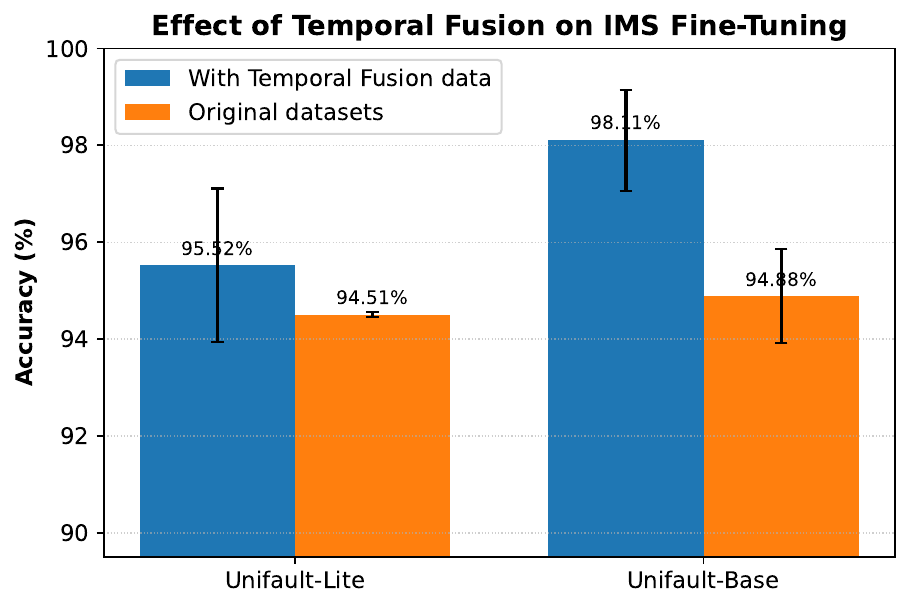}
    \includegraphics[width=0.45\columnwidth]{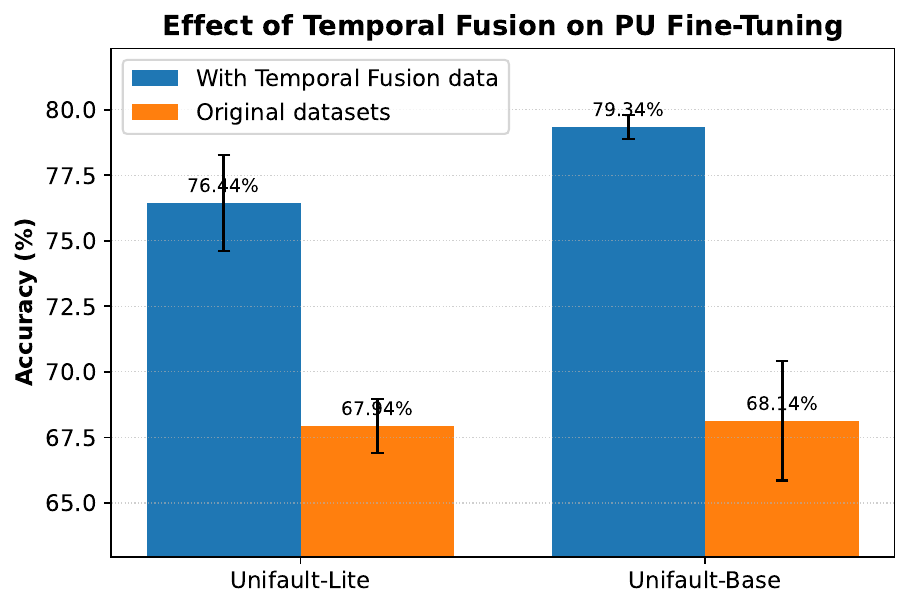}
    \caption{Effect of Cross-dataset Temporal Fusion on IMS and PU datasets for both Unifault models.}
    \label{fig:tfusion_effect}
\end{figure}

Figure~\ref{fig:tfusion_effect} shows the effect of Temporal Fusion on fine-tuning performance for the IMS and PU datasets, evaluated using \textit{Unifault-Lite} and \textit{Unifault-Base}. Temporal Fusion consistently enhances accuracy across both models and datasets, but the magnitude of improvement varies according to dataset characteristics and model capacity.

On the IMS dataset, \textit{Unifault-Lite} improves from 94.5\% to 95.5\%, while \textit{Unifault-Base} rises from 94.9\% to 98.1\%. This steady gain indicates that both models benefit from the additional temporal diversity, which helps them better capture slowly evolving fault dynamics typical of gradual bearing degradation.
For the more heterogeneous PU dataset, which includes both artificially induced and naturally degraded faults, the effect is even stronger. Accuracy increases from 67.9\% to 76.4\% for \textit{Unifault-Lite} and from 68.1\% to 79.3\% for \textit{Unifault-Base}. These results suggest that Temporal Fusion provides critical robustness to domain shifts and fault-type variability.

Overall, Temporal Fusion proves especially beneficial for complex datasets with non-stationary fault behaviors, improving generalization and stability during fine-tuning. The larger \textit{Unifault-Base} model capitalizes most on this effect, reflecting its stronger capacity to learn from temporally enriched pretraining data.

\subsubsection{Impact of Model Depth}

\begin{figure}
  \centering
  \includegraphics[width=0.7\columnwidth]{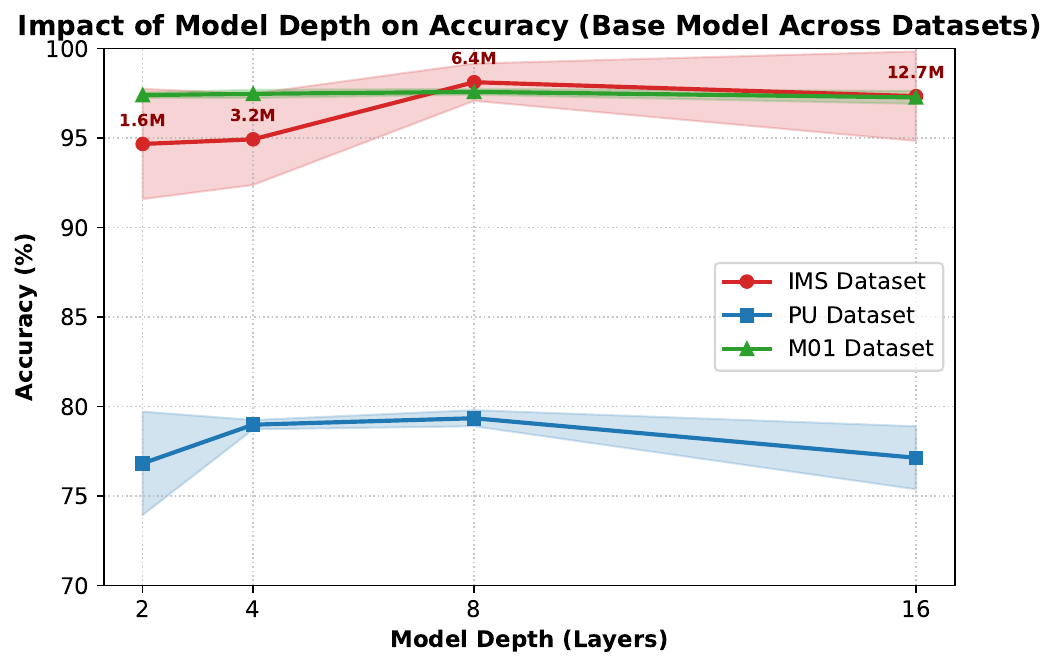}
  \caption{Effect of model depth on the Base model performance across IMS, PU, and M01 datasets (mean $\pm$ stdev over three seeds).}
  \label{fig:depth_effect}
\end{figure}
Figure~\ref{fig:depth_effect} illustrates the effect of model depth on fine-tuning accuracy for the IMS, PU, and M01 datasets using the \textit{Unifault-Base} architecture. Deeper models consistently improve accuracy on IMS and M01, while PU exhibits only moderate gains before a slight drop at higher depth.

For IMS, accuracy rises from 94.7\% at 2 layers to 98.1\% at 8 layers, confirming that additional depth enhances the model’s ability to capture long-term degradation patterns typical of naturally evolving faults. PU shows a smaller improvement, increasing from 76.8\% to 79.3\% at 8 layers before decreasing slightly to 77.1\% at 16 layers, indicating a saturation effect caused by limited data variability. The M01 dataset remains largely stable across depths, with accuracy consistently near 97.5\%, reflecting its simpler operational domain and weaker dependence on hierarchical depth.

Overall, the results suggest that moderate depth (around 8 layers) offers the best trade-off between performance and efficiency. While deeper architectures can capture complex temporal and frequency interactions, their added capacity yields diminishing returns on datasets with less temporal diversity. Thus, selecting the appropriate depth depends on the intrinsic complexity of the data, as well as the fault evolution dynamics.

\subsection{Model Analysis}
\subsubsection{Representation Geometry Before vs.\ After UniFault}

\begin{figure}
    \centering
    \setlength{\tabcolsep}{4pt}
    \begin{tabular}{cc}
        \subfloat[\textbf{CWRU} \, (Base, \emph{before} UniFault)]{
            \includegraphics[width=0.4\columnwidth]{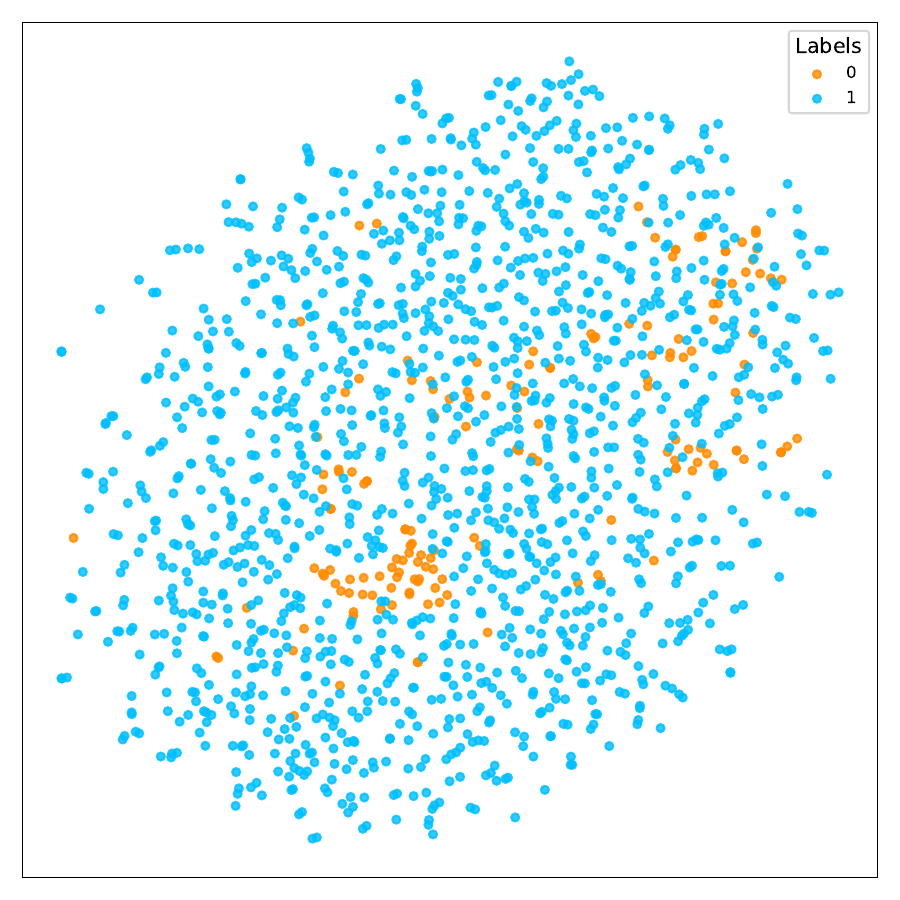}
        }
        &
        \subfloat[\textbf{CWRU} \, (Base, \emph{after} UniFault)]{
            \includegraphics[width=0.4\columnwidth]{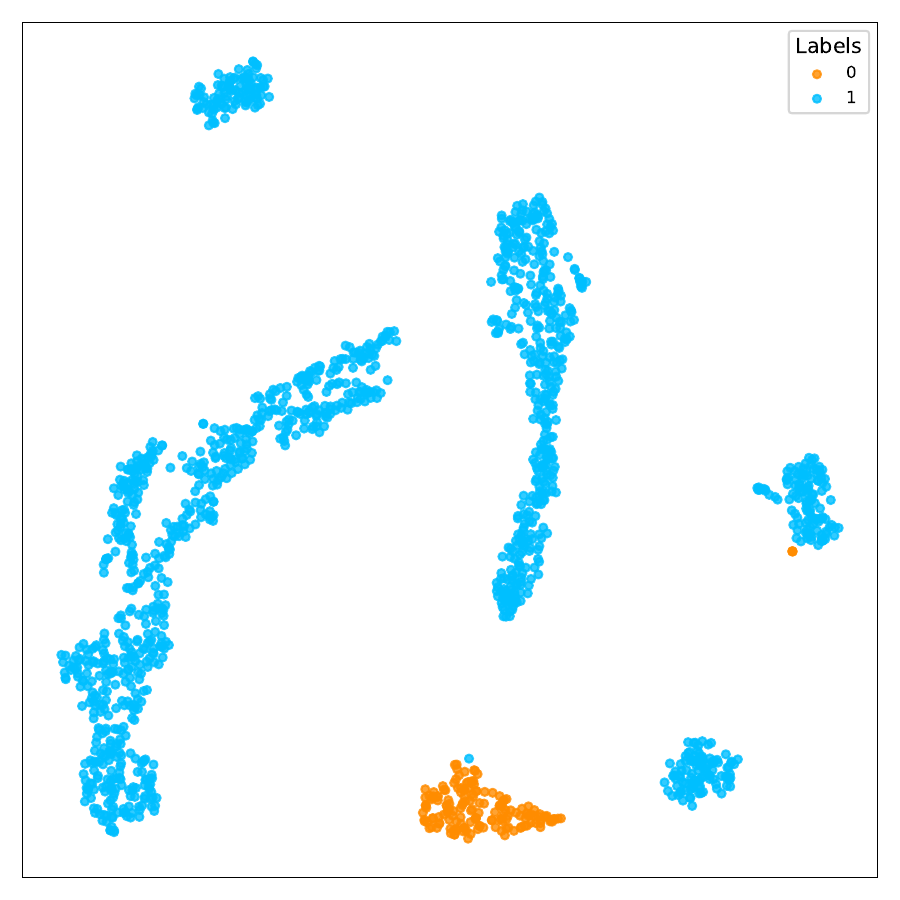}
        }
        \\
        \subfloat[\textbf{MFPT} \, (Base, \emph{before} UniFault)]{
            \includegraphics[width=0.4\columnwidth]{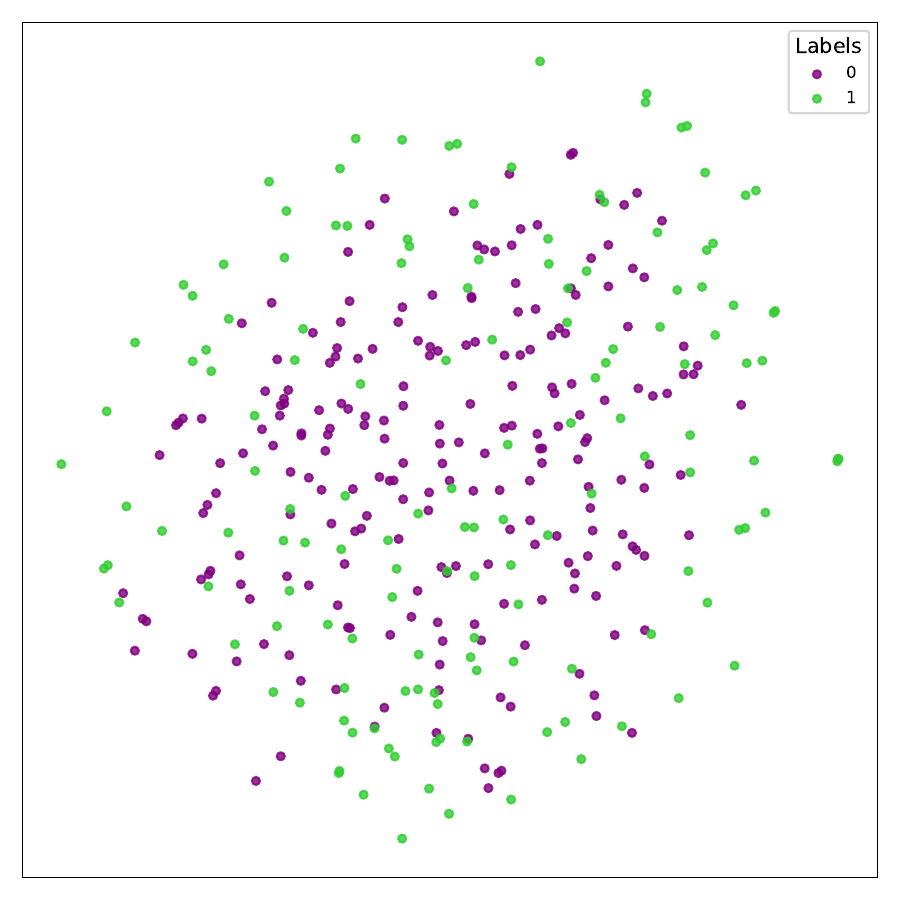}
        }
        &
        \subfloat[\textbf{MFPT} \, (Base, \emph{after} UniFault)]{
            \includegraphics[width=0.4\columnwidth]{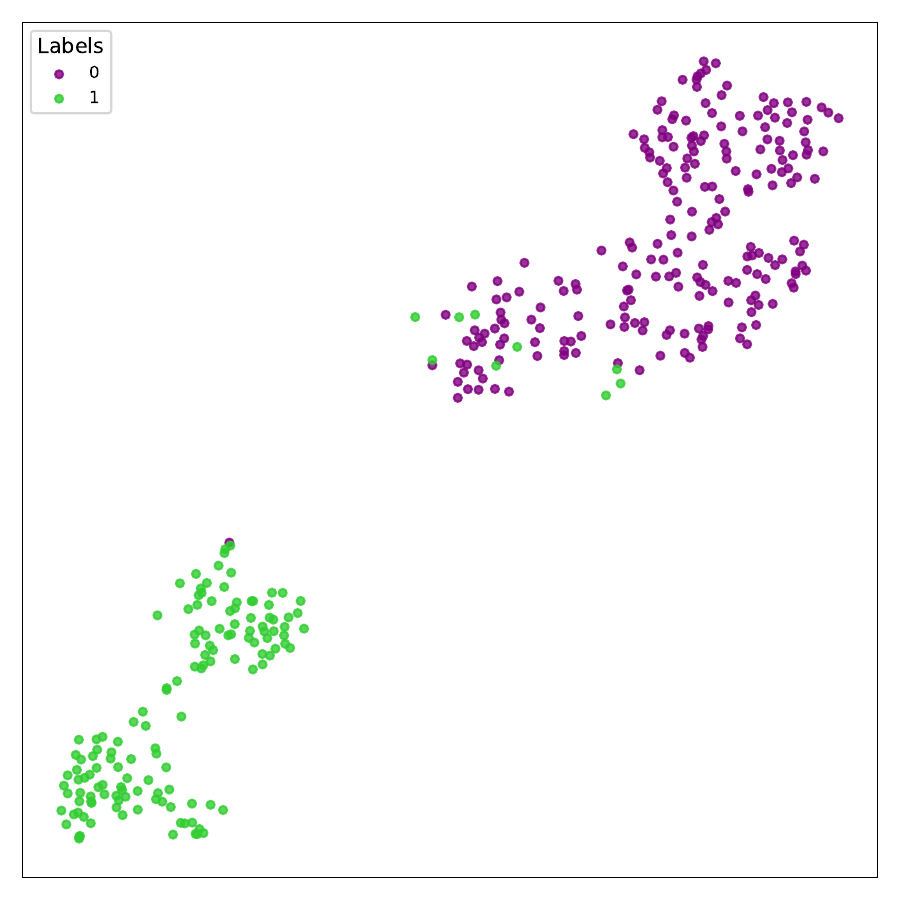}
        }
    \end{tabular}
    \caption{t\textsc{-}SNE of penultimate-layer features for CWRU and MFPT using \textit{Unifault-Base} before vs.\ after UniFault. Each point is a sample; color denotes class. UniFault tightens intra-class clusters and increases inter-class margins across both datasets.}
    \label{fig:tsne_cwru_mfpt}
\end{figure}

Figure~\ref{fig:tsne_cwru_mfpt} visualizes penultimate-layer embeddings with t\textsc{-}SNE for two datasets (CWRU and MFPT), comparing the \textit{same} model (\textit{Unifault-Base}) \textbf{before} and \textbf{after} applying UniFault. On CWRU (top row), the pre-UniFault features are fractured and partially overlapped (mixed color islands), indicating weak class separation. After UniFault, the embedding collapses into two compact, well-separated manifolds with a clear margin. We observe the same pattern on MFPT (bottom row): diffuse and interwoven clusters prior to UniFault, followed by a tight, bimodal structure afterward. Qualitatively, UniFault reduces intra-class spread while increasing inter-class separation, suggesting a cleaner decision boundary that transfers across datasets. As always with t\textsc{-}SNE, we fixed perplexity and optimization settings across all plots so differences reflect representational changes rather than visualization artifacts.

\subsubsection{K-shot Experiment}
\begin{figure}
  \centering
  \includegraphics[width=0.6\columnwidth]{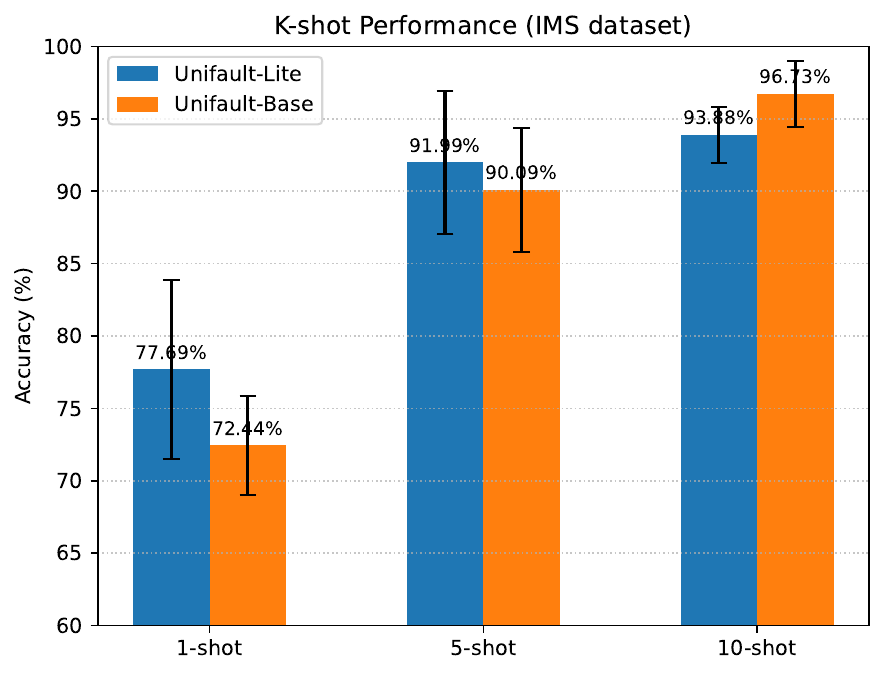}
  \caption{K-shot experiment on IMS with Unifault-Lite and Unifault-Base (mean $\pm$ std over three seeds).}
  \label{fig:few_shot_exp}
\end{figure}

Figure~\ref{fig:few_shot_exp} reports the few-shot fine-tuning results on IMS for \textit{Unifault-Lite} and \textit{Unifault-Base} using \(K\in\{1,5,10\}\) labeled examples per class. Performance improves monotonically with \(K\) for both models, and the variance decreases, indicating more stable adaptation.

At \(K=1\), \textit{Lite} attains 77.69\% while \textit{Base} reaches 72.44\%, suggesting the lighter model adapts slightly better in the extremely low-data setting. By \(K=5\), accuracies rise sharply to 91.99\% for \textit{Lite} and 90.09\% for \textit{Base}. With \(K=10\), \textit{Base} overtakes, achieving 96.73\% versus 93.88\% for \textit{Lite}. 

These results show complementary strengths: \textit{Lite} delivers stronger one-shot adaptation, whereas \textit{Base} capitalizes more on additional labeled examples, ultimately yielding the highest accuracy as \(K\) increases. In practice, the choice between the two depends on the labeling budget and latency constraints: when labels are extremely scarce, \textit{Lite} is competitive; as \(K\) grows, \textit{Base} provides superior headroom.

\subsubsection{Training Efficiency Across Model Variants}
\begin{table}[h]
    \centering
    \caption{Training Time and GPU Memory Usage Across Model Sizes. Peak GPU Memory is in GB.}
    \begin{tabular}{lccc}
        \toprule
        \textbf{Model} & \textbf{Training Time (s)} & \textbf{GPU Hours} & \textbf{Peak GPU Memory} \\
        \midrule
        Lite  & 2374  & 0.66  & 0.82  \\
        Base  & 3491  & 0.97  & 2.06  \\
        \bottomrule
    \end{tabular}
    \label{tab:training_stats}
\end{table}
To evaluate the scalability and computational demands of our proposed model, we compare the training time, GPU hours, and peak memory consumption of the three variants, as presented in Table \ref{tab:training_stats}.

The \abb-Lite variant serves as a lightweight model, requiring only 0.66 GPU hours and 0.82 GB of peak memory, making it well-suited for deployment on edge devices with constrained computational resources. The \abb-Base model offers a balance between efficiency and performance, with 0.97 GPU hours and a 2.06 GB memory footprint, making it ideal for standard industrial FD applications. 
This analysis highlights the trade-offs between model complexity and resource requirements, offering flexible deployment options depending on the target application.

\section{Conclusion and Future Work}
\label{sec:conc}
In this paper, we introduce \abb, a transformer-based foundation model for few-shot fault diagnosis in bearing data. \abb is pretrained on over 6.9 million samples from diverse, heterogeneous datasets spanning multiple applications. To address data heterogeneity, we develop a data standardization pipeline and enhance generalization by incorporating new distributions through cross-dataset temporal fusion.
Extensive evaluations across multiple datasets demonstrate that \abb consistently outperforms state-of-the-art baselines across various categories. Moreover, by pretraining different model variants, we show that \abb effectively balances performance and computational efficiency—an essential advantage for real-world deployment. Beyond achieving superior accuracy, our findings underscore the significance of scalable transformer architectures in fault diagnosis.
Overall, \abb marks a significant advancement in predictive maintenance and fault detection, offering a powerful combination of high accuracy, adaptability, and scalability, making it a strong candidate for deployment in real-world industrial applications.

Moving forward, we identify several key directions for expanding the scope and impact of our foundation model for predictive health maintenance. First, the integration of Stream-of-Quality (SoQ) methodologies will enable real-time, dynamic fault diagnosis by continuously incorporating multi-modal sensor data such as vibration and temperature \cite{LEE202258}. Embedding our foundation model into SoQ frameworks can transform static detection into adaptive, data-driven fault analytics across multi-stage manufacturing. Second, to extend beyond detection toward fault progression forecasting, we propose incorporating architectures like TimesNet  \cite{wu2023timesnet}, which are well-suited for modeling long-term, multi-scale temporal dependencies. This would enable condition-aware maintenance scheduling, allowing our model not only to detect but to anticipate future degradations. Finally, we envision our model evolving into a knowledge-driven system through integration with Industrial Large Knowledge Models (ILKM) \cite{IJAMD01624}. ILKMs can serve as a contextual backbone, linking real-time sensor streams, historical fault data, fine-tuning logs, and maintenance records. This will support automated model adaptation, systematic documentation, and interactive decision-making, enabling a fully closed-loop, intelligent diagnostic ecosystem for rotating machinery.

\bibliographystyle{unsrt}
\bibliography{references.bib}

@misc{CWRUdataset,
  title={Bearings vibration dataset. Case Western Reserve University, 2011},
  author={Loparo, KA},
  year={2015}
}

@InProceedings{pmlr-v119-zhao20c,
  title = 	 {Do {RNN} and {LSTM} have Long Memory?},
  author =       {Zhao, Jingyu and Huang, Feiqing and Lv, Jia and Duan, Yanjie and Qin, Zhen and Li, Guodong and Tian, Guangjian},
  booktitle = 	 {Proceedings of the 37th International Conference on Machine Learning},
  pages = 	 {11365--11375},
  year = 	 {2020},
  editor = 	 {III, Hal Daumé and Singh, Aarti},
  volume = 	 {119},
  series = 	 {Proceedings of Machine Learning Research},
  month = 	 {13--18 Jul},
  publisher =    {PMLR},
}

@inproceedings{PUdataset,
  title={Condition monitoring of bearing damage in electromechanical drive systems by using motor current signals of electric motors: A benchmark data set for data-driven classification},
  author={Lessmeier, Christian and Kimotho, James Kuria and Zimmer, Detmar and Sextro, Walter},
  booktitle={PHM Society European Conference},
  volume={3},
  number={1},
  year={2016}
}

@article{IMSdataset,
  title={Rexnord technical services: Bearing data set},
  author={Lee, J and Qiu, H and Yu, G and Lin, J and others},
  journal={Moffett Field, CA: IMS, Univ. Cincinnati. NASA Ames Prognostics Data Repository, NASA Ames},
  volume={12},
  pages={174102},
  year={2007}
}

@inproceedings{FEMTOdataset,
  title={PRONOSTIA: An experimental platform for bearings accelerated degradation tests.},
  author={Nectoux, Patrick and Gouriveau, Rafael and Medjaher, Kamal and Ramasso, Emmanuel and Chebel-Morello, Brigitte and Zerhouni, Noureddine and Varnier, Christophe},
  booktitle={IEEE International Conference on Prognostics and Health Management, PHM'12.},
  pages={1--8},
  year={2012},
  organization={IEEE Catalog Number: CPF12PHM-CDR}
}

@article{XJTU-SYdataset,
  title={A hybrid prognostics approach for estimating remaining useful life of rolling element bearings},
  author={Wang, Biao and Lei, Yaguo and Li, Naipeng and Li, Ningbo},
  journal={IEEE Transactions on Reliability},
  volume={69},
  number={1},
  pages={401--412},
  year={2018},
  publisher={IEEE}
}

@article{FD2015survey,
  title={A survey of fault diagnosis and fault-tolerant techniques—Part I: Fault diagnosis with model-based and signal-based approaches},
  author={Gao, Zhiwei and Cecati, Carlo and Ding, Steven X},
  journal={IEEE transactions on industrial electronics},
  volume={62},
  number={6},
  pages={3757--3767},
  year={2015},
  publisher={IEEE}
}

@article{DL_FD_survey_2019,
  title={A survey on deep learning based bearing fault diagnosis},
  author={Hoang, Duy-Tang and Kang, Hee-Jun},
  journal={Neurocomputing},
  volume={335},
  pages={327--335},
  year={2019},
  publisher={Elsevier}
}

@article{TL_FD_survey_2023,
  title={Deep transfer learning for bearing fault diagnosis: A systematic review since 2016},
  author={Chen, Xiaohan and Yang, Rui and Xue, Yihao and Huang, Mengjie and Ferrero, Roberto and Wang, Zidong},
  journal={IEEE Transactions on Instrumentation and Measurement},
  volume={72},
  pages={1--21},
  year={2023},
  publisher={IEEE}
}

@article{DA_FD_survey_2021,
  title={Applications of unsupervised deep transfer learning to intelligent fault diagnosis: A survey and comparative study},
  author={Zhao, Zhibin and Zhang, Qiyang and Yu, Xiaolei and Sun, Chuang and Wang, Shibin and Yan, Ruqiang and Chen, Xuefeng},
  journal={IEEE Transactions on Instrumentation and Measurement},
  volume={70},
  pages={1--28},
  year={2021},
  publisher={IEEE}
}

@article{DA_FD_survey_2024,
  title={Domain generalization for cross-domain fault diagnosis: An application-oriented perspective and a benchmark study},
  author={Zhao, Chao and Zio, Enrico and Shen, Weiming},
  journal={Reliability Engineering \& System Safety},
  pages={109964},
  year={2024},
  publisher={Elsevier}
}

@article{fink2020potential,
  title={Potential, challenges and future directions for deep learning in prognostics and health management applications},
  author={Fink, Olga and Wang, Qin and Svensen, Markus and Dersin, Pierre and Lee, Wan-Jui and Ducoffe, Melanie},
  journal={Engineering Applications of Artificial Intelligence},
  volume={92},
  pages={103678},
  year={2020},
  publisher={Elsevier}
}

@phdthesis{rombach2023fault,
  title={Fault Diagnostics under label and data scarcity},
  author={Rombach, Katharina},
  year={2023},
  school={ETH Zurich}
}

@article{schneider2024foundation,
  title={Foundation models: a new paradigm for artificial intelligence},
  author={Schneider, Johannes and Meske, Christian and Kuss, Pauline},
  journal={Business \& Information Systems Engineering},
  pages={1--11},
  year={2024},
  publisher={Springer}
}

@inproceedings{yuan2023power,
  title={On the power of foundation models},
  author={Yuan, Yang},
  booktitle={International Conference on Machine Learning},
  pages={40519--40530},
  year={2023},
  organization={PMLR}
}

@article{su2024machine,
  title={Machine Learning Approaches for Diagnostics and Prognostics of Industrial Systems Using Open Source Data from PHM Data Challenges: A Review},
  author={Su, Hanqi and Lee, Jay},
  journal={International Journal of Prognostics and Health Management},
  volume={15},
  number={2},
  year={2024}
}

@article{kaist_dataset,
title = {Vibration, and temperature run-to-failure dataset of ball bearing for prognostics},
journal = {Data in Brief},
volume = {54},
pages = {110403},
year = {2024},
doi = {https://doi.org/10.1016/j.dib.2024.110403},
url = {https://www.sciencedirect.com/science/article/pii/S235234092400372X},
author = {Wonho Jung and Sung-Hyun Yun and Yong-Hwa Park},
}

@article{hitsm_dataset,
doi = {10.1088/1361-6501/ac7941},
url = {https://dx.doi.org/10.1088/1361-6501/ac7941},
year = {2022},
month = {jul},
publisher = {IOP Publishing},
volume = {33},
number = {10},
pages = {105109},
author = {Wang, Zhichao and Huang, Wentao and Chen, Yi and Jiang, Yunchuan and Peng, Gaoliang},
title = {Multisource cross-domain fault diagnosis of rolling bearing based on subdomain adaptation network},
journal = {Measurement Science and Technology},
}

@article{mfpt_dataset,
title = {Intelligent bearing fault diagnosis method combining mixed input and hybrid CNN-MLP model},
journal = {Mechanical Systems and Signal Processing},
volume = {180},
pages = {109454},
year = {2022},
issn = {0888-3270},
doi = {https://doi.org/10.1016/j.ymssp.2022.109454},
url = {https://www.sciencedirect.com/science/article/pii/S0888327022005714},
author = {V. Sinitsin and O. Ibryaeva and V. Sakovskaya and V. Eremeeva},
}

@inproceedings{tstcc,
      title     = {Time-Series Representation Learning via Temporal and Contextual Contrasting},
      author    = {Eldele, Emadeldeen and Ragab, Mohamed and Chen, Zhenghua and Wu, Min and Kwoh, Chee Keong and Li, Xiaoli and Guan, Cuntai},
      booktitle = {Proceedings of the Thirtieth International Joint Conference on Artificial Intelligence, {IJCAI-21}},
      pages     = {2352--2359},
      year      = {2021}
}

@inproceedings{ts2vec,
  title={Ts2vec: Towards universal representation of time series},
  author={Yue, Zhihan and Wang, Yujing and Duan, Juanyong and Yang, Tianmeng and Huang, Congrui and Tong, Yunhai and Xu, Bixiong},
  booktitle={Proceedings of the AAAI Conference on Artificial Intelligence},
  volume={36},
  number={8},
  pages={8980--8987},
  year={2022}
}

@article{CNN_FD_review1,
  title={A comprehensive review on convolutional neural network in machine fault diagnosis},
  author={Jiao, Jinyang and Zhao, Ming and Lin, Jing and Liang, Kaixuan},
  journal={Neurocomputing},
  volume={417},
  pages={36--63},
  year={2020},
  publisher={Elsevier}
}

@article{CNN_FD1,
  title={Rolling element bearing fault diagnosis using convolutional neural network and vibration image},
  author={Hoang, Duy-Tang and Kang, Hee-Jun},
  journal={Cognitive Systems Research},
  volume={53},
  pages={42--50},
  year={2019},
  publisher={Elsevier}
}

@article{CNN_FD2,
  title={A novel multi-scale convolutional neural network incorporating multiple attention mechanisms for bearing fault diagnosis},
  author={Hu, Baoquan and Liu, Jun and Xu, Yue},
  journal={Measurement},
  volume={242},
  pages={115927},
  year={2025},
  publisher={Elsevier}
}

@article{LSTM_FD,
  title={A new hybrid LSTM-GRU model for fault diagnosis of polymer gears using vibration signals},
  author={Kumar, Anupam and Parey, Anand and Kankar, Pavan Kumar},
  journal={Journal of Vibration Engineering \& Technologies},
  volume={12},
  number={2},
  pages={2729--2741},
  year={2024},
  publisher={Springer}
}

@article{LSTM_FD2,
  title={Bi-LSTM/GRU-based anomaly diagnosis for virtual network function instance},
  author={Fan, Wentao and Yao, Jun and Cui, Shiyuan and Wang, Yan and Xu, Shuo and Tan, Yuehui and Yang, Fan and Wu, Weihong},
  journal={Computer Networks},
  volume={249},
  pages={110515},
  year={2024},
  publisher={Elsevier}
}

@article{transformer_FD1,
  title={Transformer-based intelligent fault diagnosis methods of mechanical equipment: A survey},
  author={Wang, Rongcai and Dong, Enzhi and Cheng, Zhonghua and Liu, Zichang and Jia, Xisheng},
  journal={Open Physics},
  volume={22},
  number={1},
  pages={20240015},
  year={2024},
  publisher={De Gruyter}
}

@article{transformer_FD2,
  title={Neural-transformer: A brain-inspired lightweight mechanical fault diagnosis method under noise},
  author={Wang, Changdong and Tian, Bowen and Yang, Jingli and Jie, Huamin and Chang, Yongqi and Zhao, Zhenyu},
  journal={Reliability Engineering \& System Safety},
  volume={251},
  pages={110409},
  year={2024},
  publisher={Elsevier}
}

@article{transformer_FD3,
  title={Bayesian variational transformer: A generalizable model for rotating machinery fault diagnosis},
  author={Xiao, Yiming and Shao, Haidong and Wang, Jie and Yan, Shen and Liu, Bin},
  journal={Mechanical Systems and Signal Processing},
  volume={207},
  pages={110936},
  year={2024},
  publisher={Elsevier}
}

@article{liu2024deepseek,
  title={Deepseek-v3 technical report},
  author={Liu, Aixin and Feng, Bei and Xue, Bing and Wang, Bingxuan and Wu, Bochao and Lu, Chengda and Zhao, Chenggang and Deng, Chengqi and Zhang, Chenyu and Ruan, Chong and others},
  journal={arXiv preprint arXiv:2412.19437},
  year={2024}
}

@article{team2024gemini,
  title={Gemini: A family of highly capable multimodal models, 2024},
  author={Team, Gemini and Anil, R and Borgeaud, S and Wu, Y and Alayrac, JB and Yu, J and Soricut, R and Schalkwyk, J and Dai, AM and Hauth, A and others},
  journal={arXiv preprint arXiv:2312.11805},
  year={2024}
}

@article{tcn,
  title={Wind turbine pitch bearing fault detection with Bayesian augmented temporal convolutional networks},
  author={Zhang, Chao and Zhang, Long},
  journal={Structural Health Monitoring},
  volume={23},
  number={2},
  pages={1089--1106},
  year={2024},
  publisher={SAGE Publications Sage UK: London, England}
}

@article{achiam2023gpt,
  title={Gpt-4 technical report},
  author={Achiam, Josh and Adler, Steven and Agarwal, Sandhini and Ahmad, Lama and Akkaya, Ilge and Aleman, Florencia Leoni and Almeida, Diogo and Altenschmidt, Janko and Altman, Sam and Anadkat, Shyamal and others},
  journal={arXiv preprint arXiv:2303.08774},
  year={2023}
}

@inproceedings{jin2023time,
  title={Time-LLM: Time Series Forecasting by Reprogramming Large Language Models},
  author={Jin, Ming and Wang, Shiyu and Ma, Lintao and Chu, Zhixuan and Zhang, James Y and Shi, Xiaoming and Chen, Pin-Yu and Liang, Yuxuan and Li, Yuan-Fang and Pan, Shirui and others},
  booktitle={The Twelfth International Conference on Learning Representations}
}

@inproceedings{liu2024unitime,
  title={Unitime: A language-empowered unified model for cross-domain time series forecasting},
  author={Liu, Xu and Hu, Junfeng and Li, Yuan and Diao, Shizhe and Liang, Yuxuan and Hooi, Bryan and Zimmermann, Roger},
  booktitle={Proceedings of the ACM on Web Conference 2024},
  pages={4095--4106},
  year={2024}
}

@inproceedings{liang2024foundation,
  title={Foundation models for time series analysis: A tutorial and survey},
  author={Liang, Yuxuan and Wen, Haomin and Nie, Yuqi and Jiang, Yushan and Jin, Ming and Song, Dongjin and Pan, Shirui and Wen, Qingsong},
  booktitle={Proceedings of the 30th ACM SIGKDD conference on knowledge discovery and data mining},
  pages={6555--6565},
  year={2024}
}

@ARTICLE{c-trans,
  author={Lu, Zhiqiang and Liang, Longyang and Zhu, Jun and Zou, Wenhao and Mao, Lei},
  journal={IEEE Transactions on Instrumentation and Measurement}, 
  title={Rotating Machinery Fault Diagnosis Under Multiple Working Conditions via a Time-Series Transformer Enhanced by Convolutional Neural Network}, 
  year={2023},
  volume={72},
  number={},
  pages={1-11},
  doi={10.1109/TIM.2023.3318707}
}

@article{ganin2016domain,
  author  = {Yaroslav Ganin and Evgeniya Ustinova and Hana Ajakan and Pascal Germain and Hugo Larochelle and Fran{\c{c}}ois Laviolette and Mario March and Victor Lempitsky},
  title   = {Domain-Adversarial Training of Neural Networks},
  journal = {Journal of Machine Learning Research},
  year    = {2016},
  volume  = {17},
  number  = {59},
  pages   = {1--35},
  url     = {http://jmlr.org/papers/v17/15-239.html}
}

@InProceedings{tzeng2017adversarial,
author = {Tzeng, Eric and Hoffman, Judy and Saenko, Kate and Darrell, Trevor},
title = {Adversarial Discriminative Domain Adaptation},
booktitle = {Proceedings of the IEEE Conference on Computer Vision and Pattern Recognition (CVPR)},
month = {July},
year = {2017}
}

@Article{wdcnn,
AUTHOR = {Zhang, Wei and Peng, Gaoliang and Li, Chuanhao and Chen, Yuanhang and Zhang, Zhujun},
TITLE = {A New Deep Learning Model for Fault Diagnosis with Good Anti-Noise and Domain Adaptation Ability on Raw Vibration Signals},
JOURNAL = {Sensors},
VOLUME = {17},
YEAR = {2017},
NUMBER = {2},
ARTICLE-NUMBER = {425},
URL = {https://www.mdpi.com/1424-8220/17/2/425},
PubMedID = {28241451},
ISSN = {1424-8220},
DOI = {10.3390/s17020425}
}

@ARTICLE{qcnn,
  author={Liao, Jing-Xiao and Dong, Hang-Cheng and Sun, Zhi-Qi and Sun, Jinwei and Zhang, Shiping and Fan, Feng-Lei},
  journal={IEEE Transactions on Instrumentation and Measurement}, 
  title={Attention-Embedded Quadratic Network (Qttention) for Effective and Interpretable Bearing Fault Diagnosis}, 
  year={2023},
  volume={72},
  number={},
  pages={1-13},
  doi={10.1109/TIM.2023.3259031}}

@Article{pr11020369,
    AUTHOR = {Yan, Wenhao and Wang, Jing and Lu, Shan and Zhou, Meng and Peng, Xin},
    TITLE = {A Review of Real-Time Fault Diagnosis Methods for Industrial Smart Manufacturing},
    JOURNAL = {Processes},
    VOLUME = {11},
    YEAR = {2023},
    ARTICLE-NUMBER = {369},
    URL = {https://www.mdpi.com/2227-9717/11/2/369},
    ISSN = {2227-9717},
    DOI = {10.3390/pr11020369}
}

@Article{s23115299,
AUTHOR = {Kim, Heonkook and Lee, Hojin and Kim, Seongyun and Kim, Sang Woo},
TITLE = {Attention Recurrent Neural Network-Based Severity Estimation Method for Early-Stage Fault Diagnosis in Robot Harness Cable},
JOURNAL = {Sensors},
VOLUME = {23},
YEAR = {2023},
ARTICLE-NUMBER = {5299},
URL = {https://www.mdpi.com/1424-8220/23/11/5299},
PubMedID = {37300026},
ISSN = {1424-8220},
DOI = {10.3390/s23115299}
}

@ARTICLE{8651897,
  author={Principi, Emanuele and Rossetti, Damiano and Squartini, Stefano and Piazza, Francesco},
  journal={IEEE/CAA Journal of Automatica Sinica}, 
  title={Unsupervised electric motor fault detection by using deep autoencoders}, 
  year={2019},
  volume={6},
  pages={441-451},
  doi={10.1109/JAS.2019.1911393}}

@article{guo2024domain,
  title={A domain generalization network for imbalanced machinery fault diagnosis},
  author={Guo, Yu and Ju, Guangshuo and Zhang, Jundong},
  journal={Scientific Reports},
  volume={14},
  number={1},
  pages={25447},
  year={2024},
  publisher={Nature Publishing Group UK London}
}

@article{HUANG202154,
title = {A multi-rate sampling data fusion method for fault diagnosis and its industrial applications},
journal = {Journal of Process Control},
volume = {104},
pages = {54-61},
year = {2021},
issn = {0959-1524},
doi = {https://doi.org/10.1016/j.jprocont.2021.06.003},
url = {https://www.sciencedirect.com/science/article/pii/S0959152421000949},
author = {Keke Huang and Shujie Wu and Yonggang Li and Chunhua Yang and Weihua Gui}
}

@Article{s20247065,
AUTHOR = {Pacella, Massimo and Papadia, Gabriele},
TITLE = {Fault Diagnosis by Multisensor Data: A Data-Driven Approach Based on Spectral Clustering and Pairwise Constraints},
JOURNAL = {Sensors},
VOLUME = {20},
YEAR = {2020},
ARTICLE-NUMBER = {7065},
URL = {https://www.mdpi.com/1424-8220/20/24/7065},
PubMedID = {33321733},
ISSN = {1424-8220},
DOI = {10.3390/s20247065}
}

@inproceedings{patchtst,
  title     = {A Time Series is Worth 64 Words: Long-term Forecasting with Transformers},
  author    = {Nie, Yuqi and H. Nguyen, Nam and Sinthong, Phanwadee and Kalagnanam, Jayant},
  booktitle = {International Conference on Learning Representations},
  year      = {2023}
}

@ARTICLE{cotmix,
  author={Eldele, Emadeldeen and Ragab, Mohamed and Chen, Zhenghua and Wu, Min and Kwoh, Chee-Keong and Li, Xiaoli},
  journal={IEEE Transactions on Artificial Intelligence}, 
  title={Contrastive Domain Adaptation for Time-Series Via Temporal Mixup}, 
  year={2024},
  volume={5},
  number={4},
  pages={1185-1194},
  doi={10.1109/TAI.2023.3293473}
}

@inproceedings{
vit,
title={An Image is Worth 16x16 Words: Transformers for Image Recognition at Scale},
author={Alexey Dosovitskiy and Lucas Beyer and Alexander Kolesnikov and Dirk Weissenborn and Xiaohua Zhai and Thomas Unterthiner and Mostafa Dehghani and Matthias Minderer and Georg Heigold and Sylvain Gelly and Jakob Uszkoreit and Neil Houlsby},
booktitle={International Conference on Learning Representations},
year={2021},
url={https://openreview.net/forum?id=YicbFdNTTy}
}

@article{ROCKET,
  author  = {Dempster, Angus and Petitjean, Fran\c{c}ois and Webb, Geoffrey I},
  title   = {{ROCKET}: Exceptionally Fast and Accurate Time Series Classification Using Random Convolutional Kernels},
  journal = {Data Mining and Knowledge Discovery},
  year    = {2020},
  volume  = {34},
  number  = {5},
  pages   = {1454--1495}
}

@inproceedings{onefitsall,
    title={One Fits All: Power General Time Series Analysis by Pretrained {LM}},
    author={Tian Zhou and Peisong Niu and Xue Wang and Liang Sun and Rong Jin},
    booktitle={NeurIPS},
    year={2023},
}

@inproceedings{goswami2024moment,
  title={MOMENT: A Family of Open Time-series Foundation Models},
  author={Mononito Goswami and Konrad Szafer and Arjun Choudhry and Yifu Cai and Shuo Li and Artur Dubrawski},
  booktitle={International Conference on Machine Learning},
  year={2024}
}

@article{EDWARD2025112317,
title = {EverAdapt: Continuous adaptation for dynamic machine fault diagnosis environments},
journal = {Mechanical Systems and Signal Processing},
volume = {226},
pages = {112317},
year = {2025},
issn = {0888-3270},
doi = {https://doi.org/10.1016/j.ymssp.2025.112317},
url = {https://www.sciencedirect.com/science/article/pii/S0888327025000184},
author = { Edward and Mohamed Ragab and Min Wu and Yuecong Xu and Zhenghua Chen and Abdulla Alseiari and Xiaoli Li}
}

@inproceedings{tslanet,
  title     = {TSLANet: Rethinking Transformers for Time Series Representation Learning},
  author    = {Eldele, Emadeldeen and Ragab, Mohamed and Chen, Zhenghua and Wu, Min and Li, Xiaoli},
  booktitle = {International Conference on Machine Learning},
  year      = {2024}
}

@inproceedings{anbalagan2023foundational,
  title={Foundational models for fault diagnosis of electrical motors},
  author={Anbalagan, Sriram and Agarwal, Deepesh and Natarajan, Balasubramaniam and Srinivasan, Babji},
  booktitle={2023 IEEE International Conference on Power Electronics, Smart Grid, and Renewable Energy (PESGRE)},
  pages={1--6},
  year={2023},
  organization={IEEE}
}

@article{lai2024bearingfm,
  title={BearingFM: Towards a foundation model for bearing fault diagnosis by domain knowledge and contrastive learning},
  author={Lai, Zou and Yang, Chen and Lan, Shulin and Wang, Lihui and Shen, Weiming and Zhu, Liehuang},
  journal={International Journal of Production Economics},
  volume={275},
  pages={109319},
  year={2024},
  publisher={Elsevier}
}

@article{DAGA2019252,
title = {The Politecnico di Torino rolling bearing test rig: Description and analysis of open access data},
journal = {Mechanical Systems and Signal Processing},
volume = {120},
pages = {252-273},
year = {2019},
issn = {0888-3270},
doi = {https://doi.org/10.1016/j.ymssp.2018.10.010},
url = {https://www.sciencedirect.com/science/article/pii/S0888327018306800},
author = {Alessandro Paolo Daga and Alessandro Fasana and Stefano Marchesiello and Luigi Garibaldi}
}

@article{SINITSIN2022109454,
title = {Intelligent bearing fault diagnosis method combining mixed input and hybrid CNN-MLP model},
journal = {Mechanical Systems and Signal Processing},
volume = {180},
pages = {109454},
year = {2022},
issn = {0888-3270},
doi = {https://doi.org/10.1016/j.ymssp.2022.109454},
url = {https://www.sciencedirect.com/science/article/pii/S0888327022005714},
author = {V. Sinitsin and O. Ibryaeva and V. Sakovskaya and V. Eremeeva}
}

@article{XU2023110609,
title = {A graph-guided collaborative convolutional neural network for fault diagnosis of electromechanical systems},
journal = {Mechanical Systems and Signal Processing},
volume = {200},
pages = {110609},
year = {2023},
issn = {0888-3270},
doi = {https://doi.org/10.1016/j.ymssp.2023.110609},
url = {https://www.sciencedirect.com/science/article/pii/S0888327023005174},
author = {Yadong Xu and J.C. Ji and Qing Ni and Ke Feng and Michael Beer and Hongtian Chen}
}

@article{LI2025112025,
title = {Noise-robust multi-view graph neural network for fault diagnosis of rotating machinery},
journal = {Mechanical Systems and Signal Processing},
volume = {224},
pages = {112025},
year = {2025},
issn = {0888-3270},
doi = {https://doi.org/10.1016/j.ymssp.2024.112025},
url = {https://www.sciencedirect.com/science/article/pii/S0888327024009233},
author = {Chenyang Li and Lingfei Mo and Chee Keong Kwoh and Xiaoli Li and Zhenghua Chen and Min Wu and Ruqiang Yan}
}

@article{LEE202258,
title = {Stream-of-Quality methodology for industrial Internet-based manufacturing system},
journal = {Manufacturing Letters},
volume = {34},
pages = {58-61},
year = {2022},
issn = {2213-8463},
doi = {https://doi.org/10.1016/j.mfglet.2022.09.004},
url = {https://www.sciencedirect.com/science/article/pii/S2213846322001912},
author = {Jay Lee and Prayag Gore and Xiaodong Jia and Shahin Siahpour and Pradeep Kundu and Keyi Sun}
}

@inproceedings{wu2023timesnet,
  title={TimesNet: Temporal 2D-Variation Modeling for General Time Series Analysis},
  author={Haixu Wu and Tengge Hu and Yong Liu and Hang Zhou and Jianmin Wang and Mingsheng Long},
  booktitle={International Conference on Learning Representations},
  year={2023},
}

@article{IJAMD01624,
  title={A unified industrial large knowledge model framework in Industry 4.0 and smart manufacturing},
  author={Lee, Jay and Su, Hanqi},
  journal={International Journal of AI for Materials and Design},
  issn={3041-0746},
  volume={1},
  pages={41-47},
  doi={https://doi.org/10.36922/ijamd.3681},
  url={https://accscience.com/journal/IJAMD/1/2/10.36922/ijamd.3681},
  year={2024}
}

@article{CNCdataset,
title = {Smart Data Collection System for Brownfield CNC Milling Machines: A New Benchmark Dataset for Data-Driven Machine Monitoring},
journal = {Procedia CIRP},
volume = {107},
pages = {131-136},
year = {2022},
note = {Leading manufacturing systems transformation – Proceedings of the 55th CIRP Conference on Manufacturing Systems 2022},
issn = {2212-8271},
doi = {https://doi.org/10.1016/j.procir.2022.04.022},
url = {https://www.sciencedirect.com/science/article/pii/S2212827122002384},
author = {Mohamed-Ali Tnani and Michael Feil and Klaus Diepold}
}

\end{document}